# Computer Vision Algorithm for Predicting the Welding Efficiency of Friction Stir Welded Copper Joints from its Microstructures

Akshansh Mishra[1*], Asmita Suman[2], Devarrishi Dixit[3]

[1]Department of Chemistry, Materials and Chemical Engineering "Giulio Natta", Politecnico di Milano, Milan, Italy
[2]Department of Finance, IIM Amritsar, India
[3]Department of Material Science, Christian-Albrechts-University Kiel, Germany

**Abstract:** Friction Stir Welding is a robust joining process, and numerous AI-based algorithms are being developed in this field to enhance mechanical and microstructure properties. Convolutional Neural Networks (CNNs) are Artificial Neural Networks that use image data as input. Identical to Artificial Neural Networks, they are composed of weights that are determined throughout learning, neurons (activated functions), and a goal (loss function). CNN is utilized in a variety of applications, including image recognition, semantic segmentation, image recognition, and localization. Utilizing training on 3000 microstructure pictures and new tests on 300 microstructure photographs, the current work investigates the predictions of Friction Stir Welded joint effectiveness using microstructure images.

## 1. Introduction

For scientists working with large amounts of data, machine learning has now become an indispensable tool. To process large datasets, it employs a variety of artificial intelligence (AI) techniques. It then uses the information to build digital neural networks to forecast and make decisions in various scenarios [1-5]. Outside of materials science, the forecasts can help scientists and engineers comprehend a variety of systems.

The eye of industrial automation is Computer vision. Computer vision approaches try to analyze images using cameras, sensors, and computer power in order to allow machines (human - robot interaction or other industrial instruments) to fulfill industrial activities such as production and quality assurance. Capture, process, and action are the three steps that make up the machine vision process. Computer vision is indeed a critical component of Industry 4.0. It benefits industry automation systems in a variety of ways, including increasing efficiency through better inventory management, detecting faulty items, and enhancing manufacturing quality [6-9]. Computer vision (CV) is a branch of computer science concerned with quantifying the visual contents of an image. A digital picture is a numerical expression in which each pixel is associated with a numeric gradient magnitude or a shorter vectors color value (i.e., RGB); the individual pixels convey minimal information about the image's content. The purpose of CV is to combine pixels into a high-dimensional combination of visual data to represent an image.

Many Machine Learning approaches are frequently used for the Friction Stir Welding process [10-13]. Friction Stir Welding is a robust joining procedure used to unite materials that are hard to connect using typical welding methods [14-17]. Tool Rotational Speed (RPM), Tool Traverse Rate (mm/min), as well as an Axial Force (KN) are the input variables for the Friction Stir Welding process [18-20]. Convolutional Neural Networks have been used in the Friction Stir Welding Process in only a few research. Hartl et al. [21] used Neural Network Model modeling for process monitoring with in Friction Stir Welding Process. The Fully Convolutional DenseNet-121 was used for automated visual inspection with in Friction Stir Welding process [22].

The present investigation focuses on the implementation of a Deep Convolution Network method for making predictions of the welding efficiency, that is the resilience of the weldment in reference to the resilience of the base metal of Friction Stir Welded Copper joints, based on metallographic image obtained from huge datasets available on Google and numerous other scientific papers [23-32].

## 2. Experimental Procedure

Computers process an image as numbers where each pixels of that image have some values. So an image can be represented by a set of numbers in a two-dimensional matrix. In image classification problem the main thing is to predict the class label of each images as shown in Figure 1.

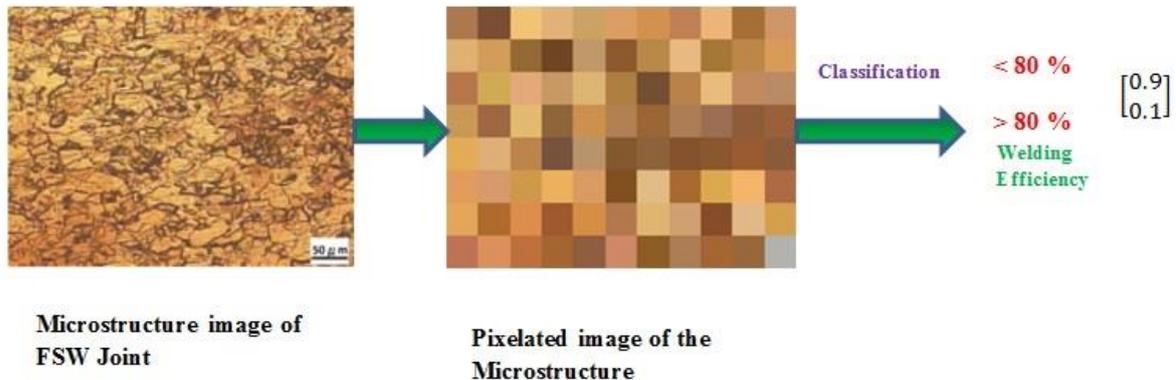

Figure 1: Classification of Microstructure images

The CNN Algorithm pipeline should be accurate enough to correctly perceive the unique features and further predict the classification of the microstructure images i.e. which microstructure image represents the welding efficiency higher than the 80 % and vice versa.

The most critical aspect of the entire machine learning process, mainly in supervised learning, is gathering a high-quality dataset. Preparing a complex dataset, on the other hand, is a lot of hard work. The first stage is to figure out what problem the model needs to answer and what kind of data may be collected. A simple end-to-end example objective may be for a car to recognize traffic signs from images captured by its own frontal camera. To generalize well and perform securely when deployed, the model would require a dataset that accurately depicts the task in as many instances as feasible. This includes various lighting and weather situations, as well as several perspectives from which the photograph could be taken. Any bias that may have been introduced inadvertently into the dataset must also be carefully considered. Otherwise, the model may discover patterns or features unrelated to traffic signs, incorrectly conclude they are significant, and make decisions based on inaccurate data.

In order to increase the number of the microstructure images, Image augmentation method is used. Data augmentation, also known as implicit regularization, is a popular strategy for improving the generalization performance of deep neural networks. It's crucial in situations where there's a scarcity of moral high-ground data and gathering additional samples is both costly and time-consuming. Image augmentation can be obtained by operations like Random Flipping, shearing, scaling etc.

Random flipping produces a 'reciprocal of the input images along one (or more) axes. Natural images may usually be turned all along horizontal axis, but not the vertical axis, because ascending and descending components of an image aren't always "interchangeable." Similarly, in this case, flipping an object by an orientation all around center pixel can be used. After that, appropriate interpolation is used to meet the original image size. Equation 1 denotes the rotation operation R, which is frequently combined with nil performed to the missing pixels.

$$R = \begin{pmatrix} \cos\alpha & -\sin\alpha \\ \sin\alpha & \cos\alpha \end{pmatrix} \qquad (1)$$

When scaled replicas of the actual microstructures images are included in the training set, the deep network can be trained on valuable deep features regardless of their original version. As demonstrated in equation 2, this process S can be done independently in multiple orientations.

$$S = \begin{pmatrix} s_x & 0 \\ 0 & s_y \end{pmatrix} \qquad (2)$$

where the scaling factors for the x and y directions are $s_x$ and $s_y$, respectively.

Each point in a microstructure image is displaced in a certain direction by the shear transformation (H). As indicated in equation 3, this offset is dependent on the distance from the line that passes through to the beginning and is transverse to this direction.

$$H = \begin{pmatrix} 1 & h_x \\ h_y & 1 \end{pmatrix} \qquad (3)$$

Where the shear coefficients in the x and y directions are denoted by $h_x$ and $h_y$, respectively.

In the present study, horizontal shift image augmentation, vertical shift image augmentation, horizontal flip image augmentation, random rotation image augmentation, brightening image augmentation, and zoom image augmentation were implemented to increase the number of microstructure dataset. The algorithm and result obtained by horizontal shift image augmentation is shown in Figure 2 and 3. The algorithm and result obtained by vertical shift image augmentation is shown in Figure 4 and 5. The algorithm and result obtained by horizontal flip image augmentation is shown in Figure 6 and 7. The algorithm and result obtained by random rotation image augmentation is shown in Figure 8 and 9. The algorithm and result obtained by brightening image augmentation is shown in Figure 10 and 11. The algorithm and result obtained by zoom image augmentation is shown in Figure 12 and 13.

```python
# example of horizontal shift image augmentation
from numpy import expand_dims
from keras.preprocessing.image import load_img
from keras.preprocessing.image import img_to_array
from keras.preprocessing.image import ImageDataGenerator
from matplotlib import pyplot
# load the image
img = load_img('img 3.png')
# convert to numpy array
data = img_to_array(img)
# expand dimension to one sample
samples = expand_dims(data, 0)
# create image data augmentation generator
datagen = ImageDataGenerator(width_shift_range=[-200,200])
# prepare iterator
it = datagen.flow(samples, batch_size=1)
# generate samples and plot
for i in range(9):
    # define subplot
    pyplot.subplot(330 + 1 + i)
    # generate batch of images
    batch = it.next()
    # convert to unsigned integers for viewing
    image = batch[0].astype('uint8')
    # plot raw pixel data
    pyplot.imshow(image)
# show the figure
pyplot.show()
```

Figure 2: Horizontal Shift Image Augmentation Algorithm

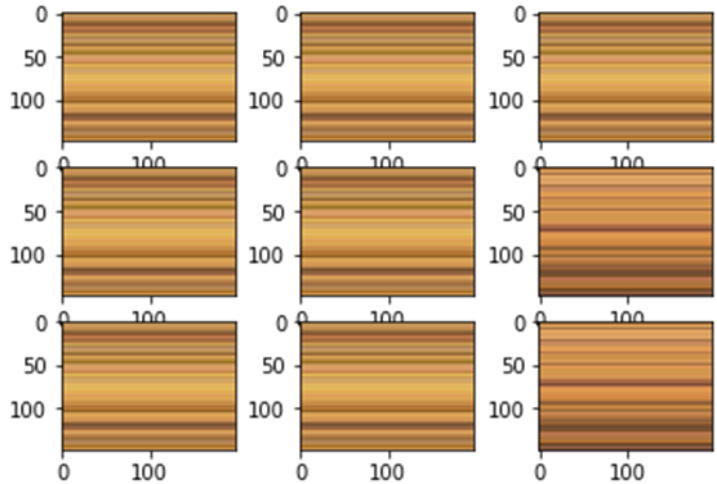

Figure 3: Augmented Microstructure images obtained by Horizontal Shift

```
# example of vertical shift image augmentation
from numpy import expand_dims
from keras.preprocessing.image import load_img
from keras.preprocessing.image import img_to_array
from keras.preprocessing.image import ImageDataGenerator
from matplotlib import pyplot
# load the image
img = load_img('img 3.png')
# convert to numpy array
data = img_to_array(img)
# expand dimension to one sample
samples = expand_dims(data, 0)
# create image data augmentation generator
datagen = ImageDataGenerator(height_shift_range=0.5)
# prepare iterator
it = datagen.flow(samples, batch_size=1)
# generate samples and plot
for i in range(9):
    # define subplot
    pyplot.subplot(330 + 1 + i)
    # generate batch of images
    batch = it.next()
    # convert to unsigned integers for viewing
    image = batch[0].astype('uint8')
    # plot raw pixel data
    pyplot.imshow(image)
# show the figure
pyplot.show()
```

Figure 4: Vertical Shift Image Augmentation Algorithm

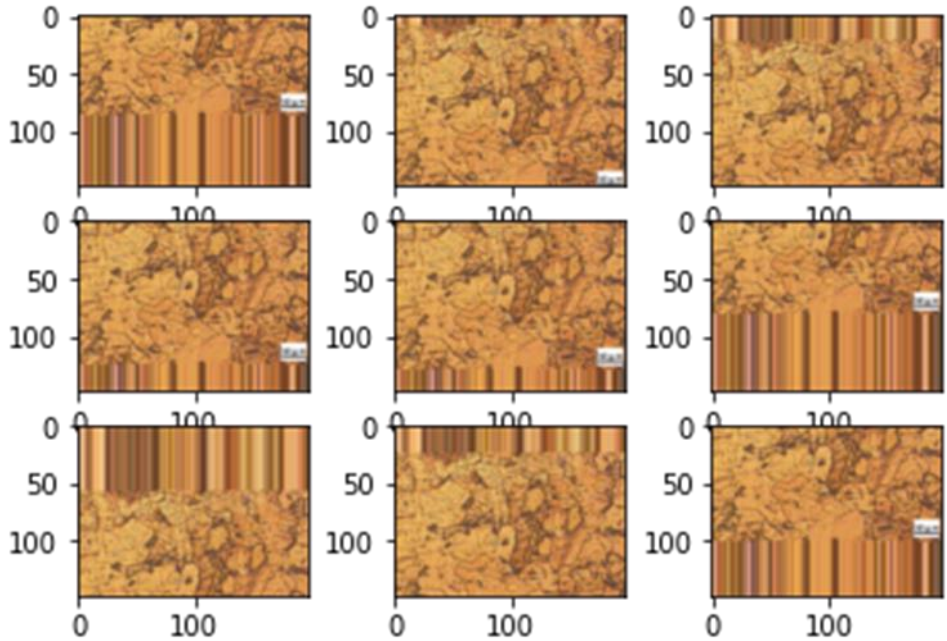

Figure 5: Augmented Microstructure images obtained by Vertical Shift

```python
# example of horizontal flip image augmentation
from numpy import expand_dims
from keras.preprocessing.image import load_img
from keras.preprocessing.image import img_to_array
from keras.preprocessing.image import ImageDataGenerator
from matplotlib import pyplot
# load the image
img = load_img('img 3.png')
# convert to numpy array
data = img_to_array(img)
# expand dimension to one sample
samples = expand_dims(data, 0)
# create image data augmentation generator
datagen = ImageDataGenerator(horizontal_flip=True)
# prepare iterator
it = datagen.flow(samples, batch_size=1)
# generate samples and plot
for i in range(9):
    # define subplot
    pyplot.subplot(330 + 1 + i)
    # generate batch of images
    batch = it.next()
    # convert to unsigned integers for viewing
    image = batch[0].astype('uint8')
    # plot raw pixel data
    pyplot.imshow(image)
# show the figure
pyplot.show()
```

Figure 6: Horizontal Flip Image Augmentation Algorithm

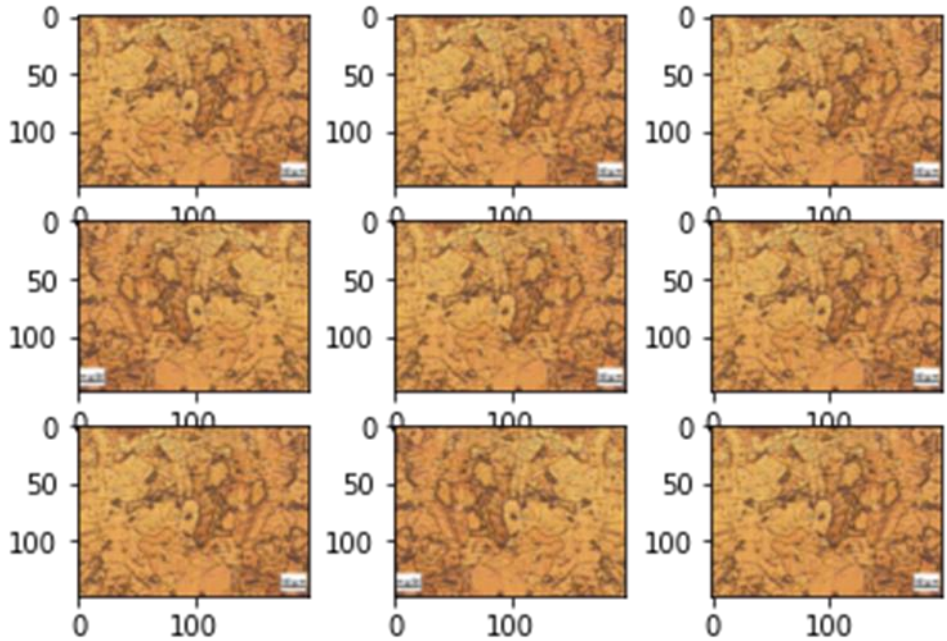

Figure 7: Augmented Microstructure images obtained by Horizontal Flip

```python
# example of random rotation image augmentation
from numpy import expand_dims
from keras.preprocessing.image import load_img
from keras.preprocessing.image import img_to_array
from keras.preprocessing.image import ImageDataGenerator
from matplotlib import pyplot
# load the image
img = load_img('img 3.png')
# convert to numpy array
data = img_to_array(img)
# expand dimension to one sample
samples = expand_dims(data, 0)
# create image data augmentation generator
datagen = ImageDataGenerator(rotation_range=90)
# prepare iterator
it = datagen.flow(samples, batch_size=1)
# generate samples and plot
for i in range(9):
    # define subplot
    pyplot.subplot(330 + 1 + i)
    # generate batch of images
    batch = it.next()
    # convert to unsigned integers for viewing
    image = batch[0].astype('uint8')
    # plot raw pixel data
    pyplot.imshow(image)
# show the figure
pyplot.show()
```

Figure 8: Random Rotation Image Augmentation Algorithm

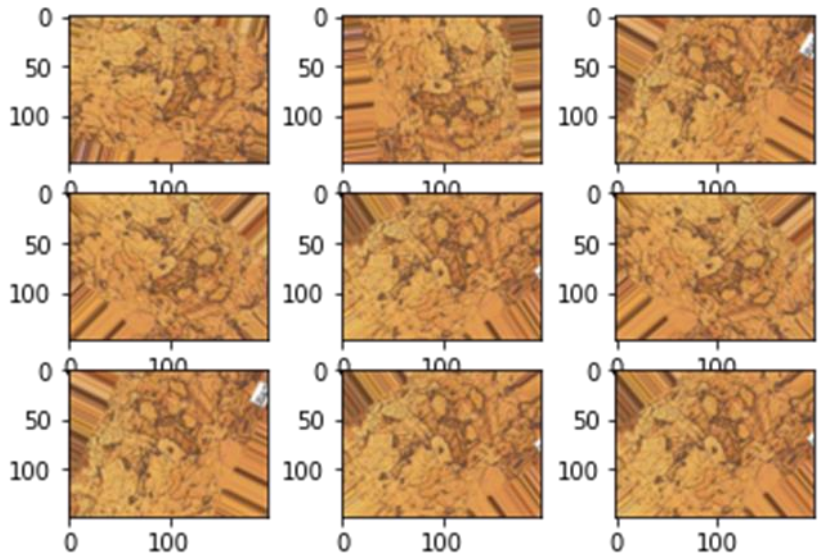

Figure 9: Augmented Microstructure images obtained by Random Rotation

```python
# example of brighting image augmentation
from numpy import expand_dims
from keras.preprocessing.image import load_img
from keras.preprocessing.image import img_to_array
from keras.preprocessing.image import ImageDataGenerator
from matplotlib import pyplot
# load the image
img = load_img('img 3.png')
# convert to numpy array
data = img_to_array(img)
# expand dimension to one sample
samples = expand_dims(data, 0)
# create image data augmentation generator
datagen = ImageDataGenerator(brightness_range=[0.2,1.0])
# prepare iterator
it = datagen.flow(samples, batch_size=1)
# generate samples and plot
for i in range(9):
    # define subplot
    pyplot.subplot(330 + 1 + i)
    # generate batch of images
    batch = it.next()
    # convert to unsigned integers for viewing
    image = batch[0].astype('uint8')
    # plot raw pixel data
    pyplot.imshow(image)
# show the figure
pyplot.show()
```

Figure 10: Brightening Image Augmentation Algorithm

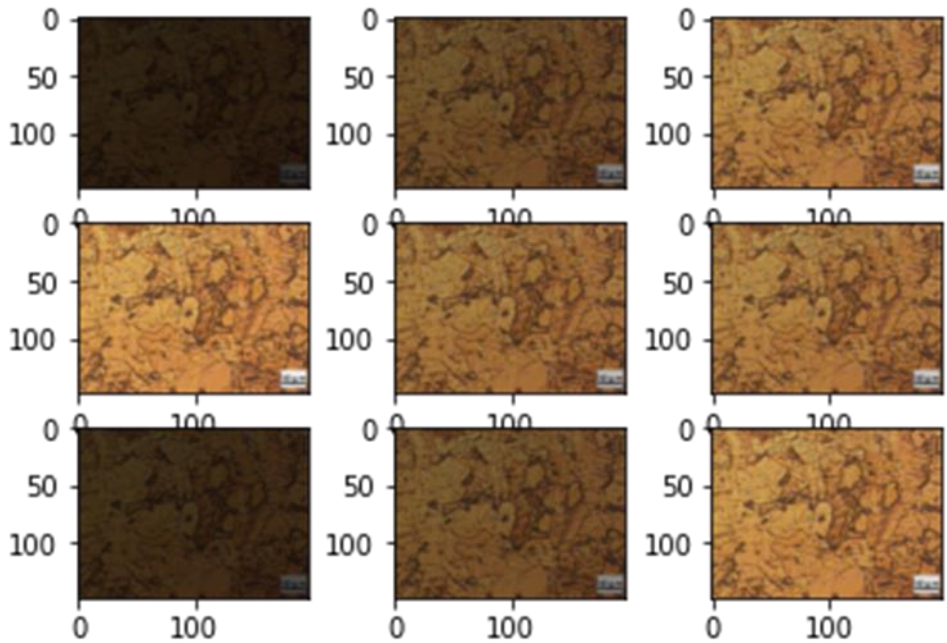

Figure 11: Augmented Microstructure images obtained by Brightening

```python
# example of zoom image augmentation
from numpy import expand_dims
from keras.preprocessing.image import load_img
from keras.preprocessing.image import img_to_array
from keras.preprocessing.image import ImageDataGenerator
from matplotlib import pyplot
# load the image
img = load_img('img 3.png')
# convert to numpy array
data = img_to_array(img)
# expand dimension to one sample
samples = expand_dims(data, 0)
# create image data augmentation generator
datagen = ImageDataGenerator(zoom_range=[0.5,1.0])
# prepare iterator
it = datagen.flow(samples, batch_size=1)
# generate samples and plot
for i in range(9):
    # define subplot
    pyplot.subplot(330 + 1 + i)
    # generate batch of images
    batch = it.next()
    # convert to unsigned integers for viewing
    image = batch[0].astype('uint8')
    # plot raw pixel data
    pyplot.imshow(image)
# show the figure
pyplot.show()
```

Figure 12: Zoom Image Augmentation Algorithm

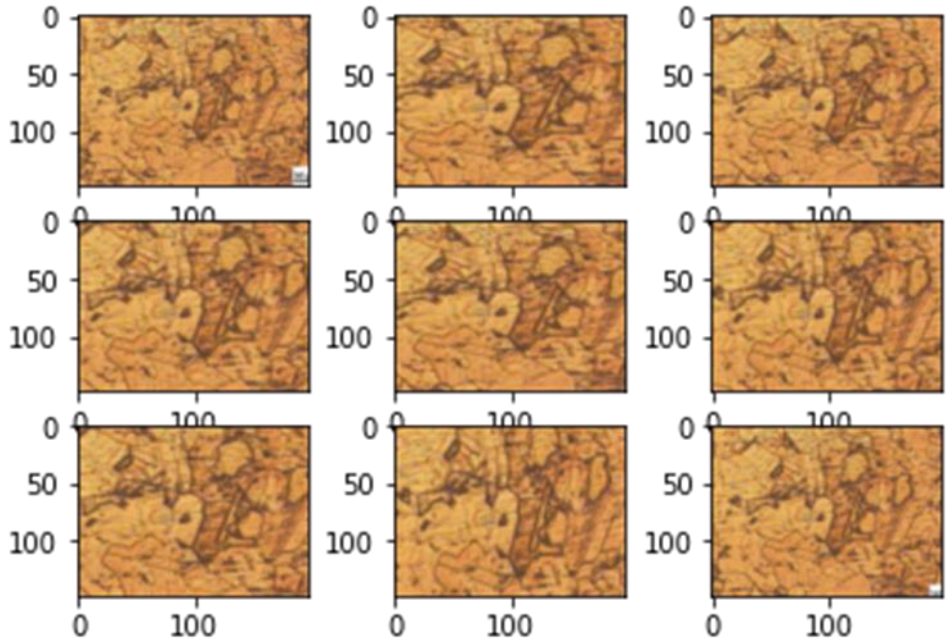

Figure 13: Augmented Microstructure images obtained by Zooming

In Computer Vision, datasets are divided into two types: training datasets and testing datasets. Training datasets are used to teach the algorithm to do the desired task, while testing datasets are used to test the approach. The microstructure pictures with welding efficiency just under 80% and microstructure imaging with welding efficiency more than or equal to 80% were used in this investigation.

## 3. Results and Discussion

The main goal of the CNN architecture is to learn the features directly from the data. There are three main parts of the CNN architecture i.e. Convolution, Non-linearity and Pooling as shown in Figure 14.

The main function of the convolution is to extract the features from the microstructure image or from the previous layer. Non-linearity is introduced to deal with the non-linear data and further introduce the complexity in the learning pipeline in order to solve more complex tasks. Pooling operation allows to down sample the spatial resolution of the microstructure image or multiple scale features of the microstructure image. Computation of the class scores can be outputted by a dense layer which is after the convolution layer.

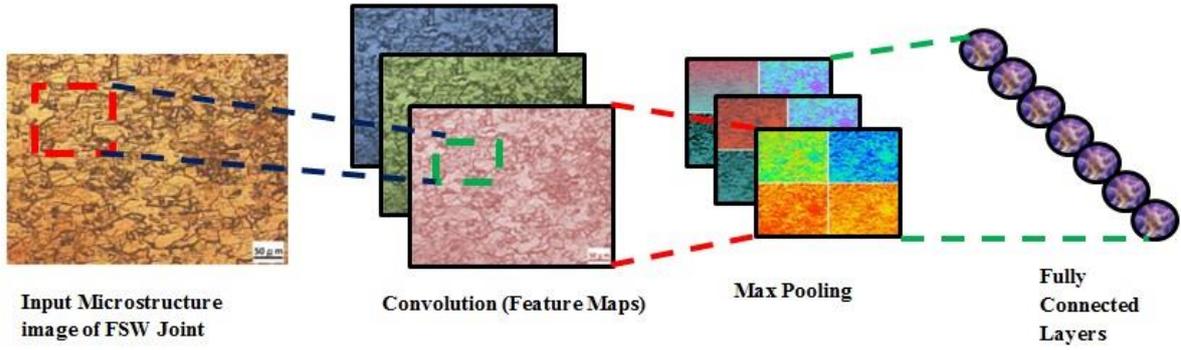

Figure 14: Representation of the CNN Architecture

Each neuron in the hidden layer computes a weighted sum of the inputs from that patch of an input microstructure image as shown in Figure and further bias is applied which is activated by a local non-linearity. The actual computation for a neuron in that hidden layer is defined by equation 4.

$$\sum_{i=1}^{4}\sum_{j=1}^{4} w_{ij} x_{i+p, j+q} + b \qquad (4)$$

Where w is the weight, x is the input features and b is the bias.

Consider the following scenario: if we have microstructure image and we want to perform an image pattern classification operation. The input is a microstructure of a friction stir fused joint, and we want to know whether it belongs to the above 80 percent welding efficiency or below 80 percent welding efficiency classification. Let's we have have a 40 x 40 x 3 image as shown in Figure 15.

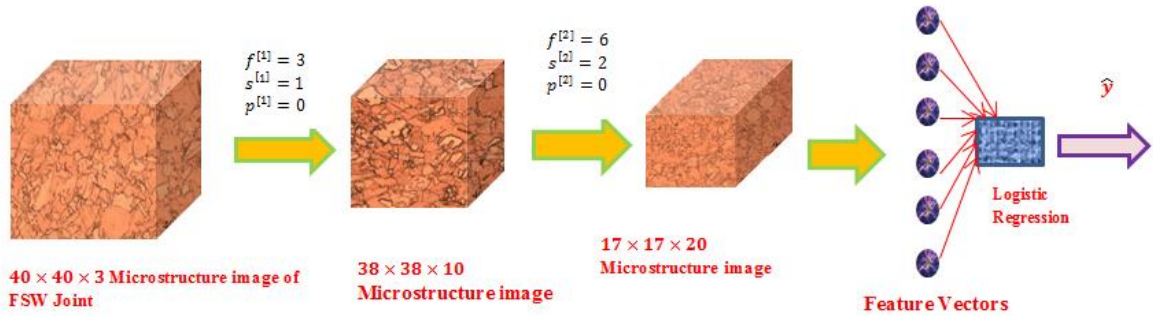

Figure 15: Convolution operation on a given microstructure image

Height and width of the microstructure image can be represented by $n_H^{[0]} = n_W^{[0]} = 40$ and number of channels is represented by $n_C^{[0]} = 3$. The first layer uses 3 X3 filters to detect the features which are represented by $f^{[1]} = 3$ and there is a stride of 1 with no padding. By using ten convolutions it is assumed that there are 10 filters then the activations in the next layer of a neural network will be 38 X 38 X 10. The new value of $n_H{}^{[1]}$ and $n_W{}^{[1]}$ is calculated by Equation 5.

$$n_H{}^{[1]} = \frac{n+2p-f}{s} + 1 \qquad (5)$$

So the new values for height and width will be $n_H{}^{[1]} = n_W{}^{[1]} = 38$ and new value for a number of channels is represented by $n_C{}^{[1]} = 10$. These become the dimensions of the activations at a first layer. Now let's say we have n another convolutional layer and 6 X 6 filters are used represented by $f^{[2]} = 6$ having a stride of $s^{[2]} = 2$ with no padding and number of filters is 20. It is observed that the dimension had shrunk much faster to 17 X 17 X 20. So overall it is observed that we have taken 40 X 40 X 3 input image and computed it to 17 X 17 X 20 =5780 features for the microstructure image and what's commonly done is flattening into vectors and unrolling the volume of the final microstructure image into 5780 units which is further fed to a logistic regression unit or a soft-max unit to recognize whether the microstructure images belong to the welding efficiency of greater than 80 % or to a welding efficiency less than 80 % and it further yields the output $\hat{y}$. Figure 16 shows the CNN architecture used in the present study.

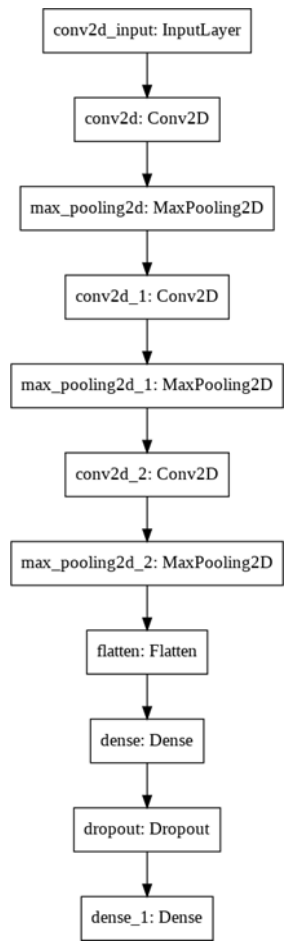

Figure 16: CNN Architecture used in the present study

The plot of the loss function against the number of epochs is shown in Figure 17. The plot of accuracy against the number of epochs is shown in Figure 18. The accuracy increases for both the training and testing (validation) datasets, as shown in Figure 18.

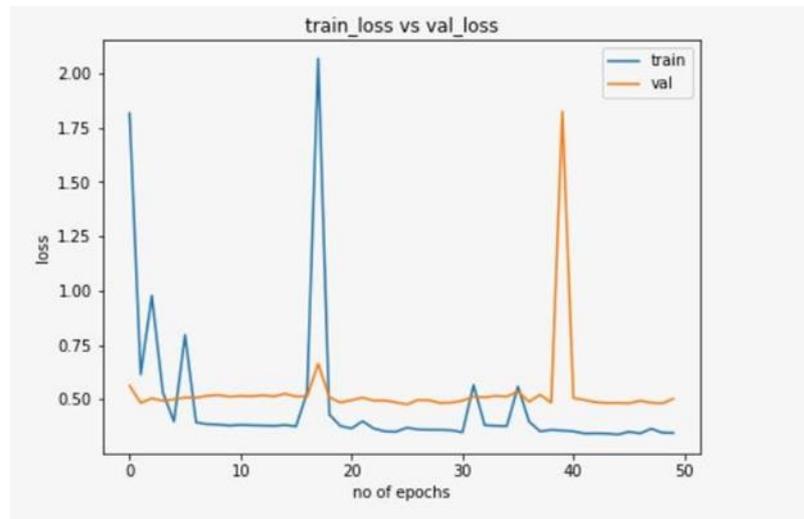

Figure 17: Loss function plotted against the number of epochs

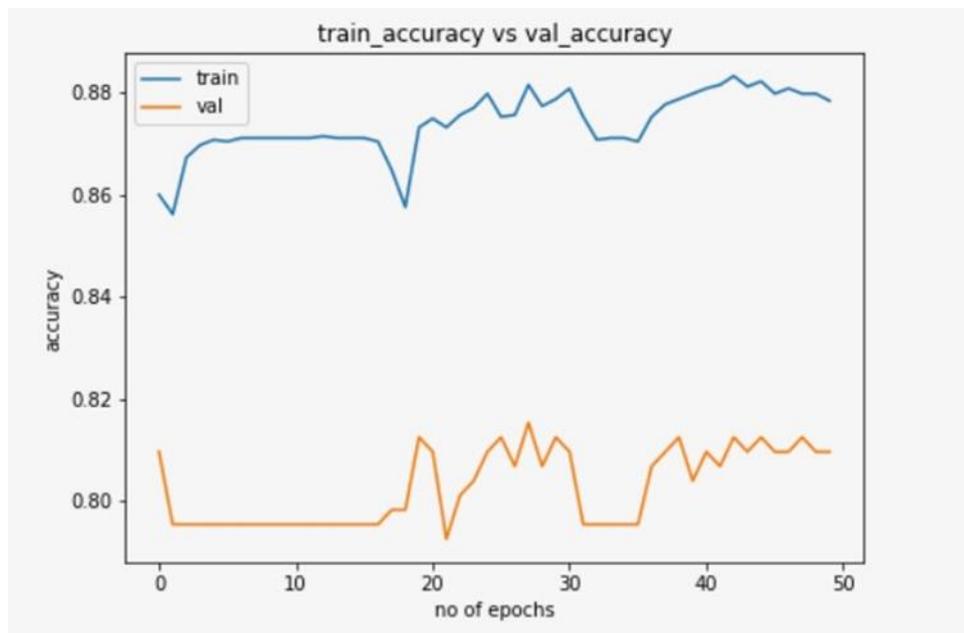

Figure 18: Accuracy as a function of the number of epochs

## 4. Conclusion

In this study, a Deep Convolutional Neural Network is effectively used to predict the welding effectiveness of Friction Stir Welded connections, which is a mechanical feature. Convolutional Neural Networks (CNNs) are a strong tool for image identification and sophisticated problem solving. A few transforms should be kept in mind when training

Artificial Neural Network models. For binary classification issues, there is a topmost layer with a nonlinear activation function as well as a binary node. The current research project is primarily focused on developing a basic Classification algorithm for classifying the two types of microstructures. The model was created with such an accuracy of 80%, however it may be better. The model may be improved further by:

- including more photographs into the dataset
- modifying the learning rate
- altering the batch size, which will help the system to recognize patterns more accurately. When the batch size is small, the patterns repeat less frequently, making convergence difficult, whereas when the batches size is large, learning is slow.
- Increasing or lowering the amount of epochs

The effort will continue to employ the Fully Convolutional (CNN) technique to detect flaws in Friction Stir Welded joints in the future.

*References*

1. Russell, S. and Norvig, P., 2002. Artificial intelligence: a modern approach.

2. Nilsson, N.J., 2009. The quest for artificial intelligence. Cambridge University Press.

3. Minsky, M., 1961. Steps toward artificial intelligence. Proceedings of the IRE, 49(1), pp.8-30.

4. Negnevitsky, M., 2005. Artificial intelligence: a guide to intelligent systems. Pearson education.

5. Cohen, P.R. and Feigenbaum, E.A. eds., 2014. The Handbook of Artificial Intelligence: Volume 3 (Vol. 3). Butterworth-Heinemann.

6. Jarvis, R.A., 1983. A perspective on range finding techniques for computer vision. IEEE Transactions on Pattern Analysis and Machine Intelligence, (2), pp.122-139.

7. Voulodimos, A., Doulamis, N., Doulamis, A. and Protopapadakis, E., 2018. Deep learning for computer vision: A brief review. Computational intelligence and neuroscience, 2018.

8. Jähne, B., Haussecker, H. and Geissler, P. eds., 1999. Handbook of computer vision and applications (Vol. 2, pp. 423-450). New York: Academic press.

9. Faugeras, O. and Faugeras, O.A., 1993. Three-dimensional computer vision: a geometric viewpoint. MIT press.

10. Du, Y., Mukherjee, T. & DebRoy, T. Conditions for void formation in friction stir welding from machine learning. npj Comput Mater 5, 68 (2019). https://doi.org/10.1038/s41524-019-0207-y

11. Du, Y., Mukherjee, T., Mitra, P. and DebRoy, T., 2020. Machine learning based hierarchy of causative variables for tool failure in friction stir welding. Acta Materialia, 192, pp.67-77.

12. Mishra, A. and Patti, A., 2021. Deep Convolutional Neural Network Modeling and Laplace Transformation Algorithm for the Analysis of Surface Quality of Friction Stir Welded Joints.

13. Mishra, A. and Pathak, T., 2020. Estimation of Grain Size Distribution of Friction Stir Welded Joint by using Machine Learning Approach.

14. Mishra, R.S. and Ma, Z.Y., 2005. Friction stir welding and processing. Materials science and engineering: R: reports, 50(1-2), pp.1-78.

15. Rai, R., De, A., Bhadeshia, H.K.D.H. and DebRoy, T., 2011. friction stir welding tools. Science and Technology of welding and Joining, 16(4), pp.325-342.

16. Lohwasser, D. and Chen, Z. eds., 2009. Friction stir welding: From basics to applications. Elsevier.

17. Thomas, W.M., Johnson, K.I. and Wiesner, C.S., 2003. Friction stir welding–recent developments in tool and process technologies. Advanced engineering materials, 5(7), pp.485-490.

18. Gite, R.A., Loharkar, P.K. and Shimpi, R., 2019. Friction stir welding parameters and application: A review. Materials Today: Proceedings, 19, pp.361-365.

19. Mishra, A., 2018. Friction stir welding of dissimilar metal: a review. Available at SSRN 3104223.

20. Mishra, A. and Dixit, D., 2018. Friction Stir Welding of Aerospace Alloys. Journal of Mechanical Engineering, 48(1), pp.37-46.

21. Hartl, R.; Bachmann, A.; Habedank, J.B.; Semm, T.; Zaeh, M.F. Process Monitoring in Friction Stir Welding Using Convolutional Neural Networks. Metals 2021, 11, 535. https://doi.org/10.3390/met11040535

22. R. Hartl, J. Landgraf, J. Spahl, A. Bachmann, and M. F. Zaeh "Automated visual inspection of friction stir welds: a deep learning approach", Proc. SPIE 11059, Multimodal Sensing: Technologies and Applications, 1105909 (21 June 2019); https://doi.org/10.1117/12.2525947

23. Sakthivel, T. and Mukhopadhyay, J., 2007. Microstructure and mechanical properties of friction stir welded copper. Journal of Materials Science, 42(19), pp.8126-8129.

24. Lee, W.B. and Jung, S.B., 2004. The joint properties of copper by friction stir welding. Materials Letters, 58(6), pp.1041-1046.

25. Mironov, S., Inagaki, K., Sato, Y.S. and Kokawa, H., 2015. Microstructural evolution of pure copper during friction-stir welding. Philosophical Magazine, 95(4), pp.367-381.

26. Savolainen, K., 2012. Friction stir welding of copper and microstructure and properties of the welds.

27. Sun, Y.F. and Fujii, H., 2010. Investigation of the welding parameter dependent microstructure and mechanical properties of friction stir welded pure copper. Materials Science and Engineering: A, 527(26), pp.6879-6886.

28. Shen, J.J., Liu, H.J. and Cui, F., 2010. Effect of welding speed on microstructure and mechanical properties of friction stir welded copper. Materials & Design, 31(8), pp.3937-3942.

29. Kumar, A. and Raju, L.S., 2012. Influence of tool pin profiles on friction stir welding of copper. Materials and Manufacturing Processes, 27(12), pp.1414-1418.

30. Xie, G.M., Ma, Z.Y. and Geng, L., 2007. Development of a fine-grained microstructure and the properties of a nugget zone in friction stir welded pure copper. Scripta Materialia, 57(2), pp.73-76.

31. Liu, H.J., Shen, J.J., Huang, Y.X., Kuang, L.Y., Liu, C. and Li, C., 2009. Effect of tool rotation rate on microstructure and mechanical properties of friction stir welded copper. Science and Technology of welding and Joining, 14(6), pp.577-583.

32. Sun, Y.F. and Fujii, H., 2011. The effect of SiC particles on the microstructure and mechanical properties of friction stir welded pure copper joints. Materials Science and Engineering: A, 528(16-17), pp.5470-5475.